# Face Verification Using Kernel Principle Component Analysis

| Manjupriya, | Chithra. C .K | Divya. M | Karthikeyan.V |
|---|---|---|---|
| *U.G Student,* | *U.G Student,* | *U.G Student,* | *Assistant Professor,* |
| *Dept. of ECE* | *Dept. of ECE* | *Dept. of ECE* | *Dept. of ECE* |

**ABSTRACT**

*In the beginning stage, face verification is done using easy method of geometric algorithm models, but the verification route has now developed into a scientific progress of complicated geometric representation and matching process. In modern time the skill have enhanced face detection system into the vigorous focal point. Researcher's currently undergoing strong research on finding face recognition system for wider area information taken under hysterical elucidation dissimilarity. The proposed face recognition system consists of a narrative exposition-indiscreet preprocessing method, a hybrid Fourier-based facial feature extraction and a score fusion scheme. We take in conventional the face detection in unlike cheer up circumstances and at unusual setting. Image processing, Image detection, Feature- removal and Face detection are the methods used for Face Verification System (FVS). This paper focuses mainly on the issue of toughness to lighting variations. The proposed system has obtained an average of 81.5% verification rate on Two-Dimensional images under different lightening conditions.*

## 1.INTRODUCTION

In the past decades, many appearance-based methods have been proposed to handle this problem, and new theoretical insights as well as good recognition results have been reported. In the proposed the verification of the face in different climatic conditions, this paper focuses mainly on the issue of robustness to lighting variations. Established advance for production with this concern can be broadly classified into three types based on look, normalized conditions, and qualitymethods. In direct appearance-based approaches, training examples are collected under different lighting conditions and directly (i.e., without undergoing any lighting preprocessing) used to learn a global model of the possible illumination variations but it requires a large number of training images and an expressive feature set, otherwise it is essential to include a good preprocessor to reduce illumination variations. In this paper verify the face in different lightening conditions means any time (day or night) and in any place (indoor or outdoor). There are six methods used to improve the face verification rate of image. Preprocessing chain, Local ternary patterns, Local binary patterns, Gabor wavelet, and phase congruency are the methods of face verification system. This paper focuses mainly on the issue of robustness to lighting variations. Traditional approaches for dealing with this issue can be broadly classified into three categories: appearance-based, normalization-based, and feature-based methods. In direct appearance-based approaches, training examples are collected under different lighting conditions and directly (i.e., without undergoing any lighting preprocessing) used to learn a global model of the possible illumination variations but it requires a large number of training images and an expressive feature set, otherwise it is essential to include a good preprocessor to reduce illumination variations. Normalization based approaches seek to reduce the image to a more "canonical" form in which the illumination variations are suppressed. Histogram equalization is one simple example of this method. These methods are quite effective but their ability to handle spatially non uniform variations remains limited. The third approach extracts illumination-insensitive feature sets directly from the given image. These feature sets range from geometrical features to image derivative features such as edge maps, local binary patterns (LBP), Gabor wavelets, and local Autocorrelation filters. Although such features offer a great improvement on raw gray values, their resistance to the complex illumination variations that occur in real-world face images is still quite limited.

The integrative framework is proposed, that combines the strengths of all three of the above approaches.

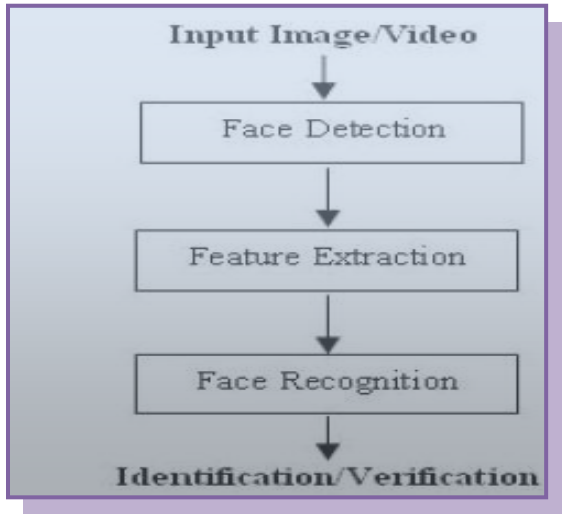

**Figure 1. Three major parts of face recognition algorithm**

The overall process can be viewed as a pipeline consisting of image normalization, feature extraction, and subspace representation, as shown. Each stage increases resistance to illumination variations and makes the information needed for recognition more manifest. This method achieves very significant improvements, than the other method of verification rate is 88.1% at 0.1% false acceptance rate.

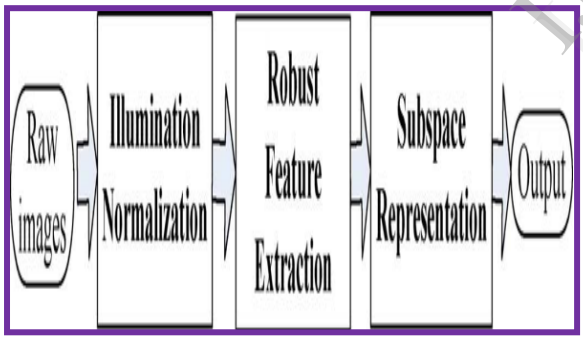

Figure .2 Stages of our full face recognition method

Several aspects of the relationship between image normalization and feature sets, Robust feature sets and feature comparison strategies, Fusion of multiple feature sets framework is the will be verified

## 2. Principal Components Analysis PCA)

PCA commonly referred to as the use of eigen faces, is the technique pioneered by Kirby and Sirivich. With PCA, the probe and gallery images must be the same size and must first be normalized to line up the eyes and mouth of the subjects within the image. The PCA approach is then used to reduce the dimension of the data by means of data compression basics and reveals the most effective low dimensional structure of facial patterns. This reduction in dimensions removes information that is not useful and precisely decomposes the face structure into orthogonal components known as Eigen faces. Each face image may be represented as the weighted sum of the Eigen faces, which are stored in a 1D array. A probe image is compared against a gallery image by measuring the distance between their respective feature vectors. The PCA approach typically requires the full frontal face to be presented each time; otherwise the image results in poor performance. The primary advantage of this technique is that it can reduce the data needed to identify the individual to $1/1000^{th}$ of the data presented.

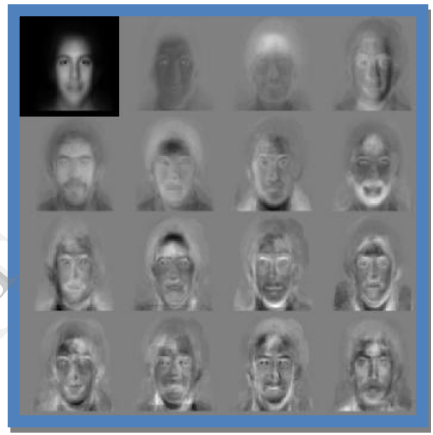

Fig .3: Standard Eigen faces: Feature vectors are derived using Eigen faces.

### 2.1 Linear Discriminant Analysis (LDA)

LDA is a statistical approach for classifying samples of unknown classes based on training samples with classes. This technique aims to maximize between-class variance and minimize within –class variance. In fig , where each block represents a class, there are large variances between classes, but little variance within classes. When dealing with high dimensional face data, this technique faces the small sample size problem that arises where there are a small number of available training samples compared to the dimensionality of the sample space.

## 3. Existing System

Several face recognition techniques have been introduced to identify the face in difficult lighting condition. Face recognition has received a great deal of attention from the scientific and industrial communities over the past several decade s owing to its wide range of applications in information security and access control, law enforcement, surveillance, and more generally image understanding.

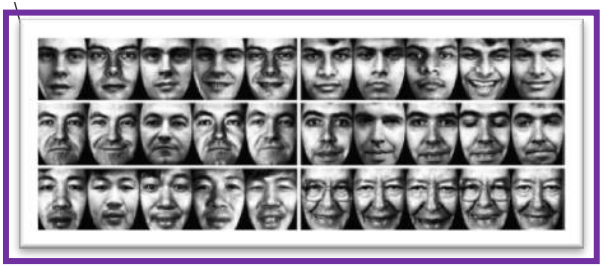

Fig .4: Examples of six classes using LDA

### 3.1 EIGEN FACES

The Eigen face algorithm is proved to be very successful in face recognition the Eigen face method is also based on linearly projecting the image space to a low dimensional feature space. However, the Eigen face method, which uses principal components analysis (PCA) for dimensionality reduction, yields projection directions that maximize the total scatter across all classes, i.e., across all images of all faces. In choosing the projection which maximizes total scatter, PCA retains unwanted variations due to lighting and facial expression. As correlation methods are computationally expensive and require great amounts of storage, it is natural to pursue dimensionality reduction schemes. A technique now commonly used for dimensionality reduction in computer vision particularly in face recognition is principal components analysis (PCA). PCA techniques, also known as Karhunen-Loeve methods, choose a dimensionality reducing linear projection that maximizes the scatter of all projected samples. This method uses the face pictures as raw data and extracts some feature vectors from these pictures. So the classifier can be trained with these feature vectors. If the illumination conditions of the environment changes, then the efficiency of the system will decrease because the value of the pixels in the test image will change significantly. This is a big problem with the eigen face method.

We can see in that the edge information in eigen face pictures is not destroyed. If the edge information of face is applied in eigen face method, instead of the face image, we anticipate that the illumination dependency problem of eigen face method will be resolved. This means that we want to extract eigenvectors from edge information of faces and not from face images. The advantage of this idea, if can be implemented, is the robustness of eigen face method under varying illumination conditions.

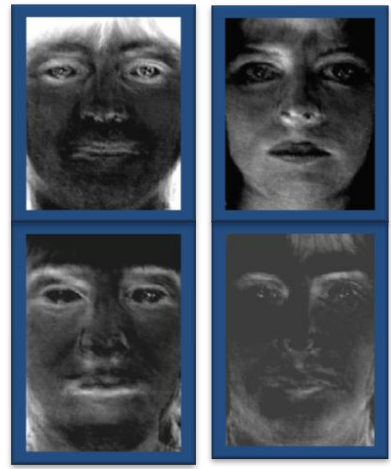

Figure .5 four first Eigen faces of a training set of 23 images.

A drawback of this approach is that the scatter being maximized is due not only to the between-class scatter that is useful for classification, but also to the within-class scatter that, for classification purposes, is unwanted information.

### 3.2 LINEAR DISCRIMINANT ANALYSIS

LDA is a supervised learning method that finds the linear projection in subspaces. It maximizes the between-class scatter while minimizing the within-class scatter of the projected data According to this objective, two scatter matrices—the between class Scatter $S_B$ matrix and the within-class scatter matrices $S_W$ are defined as

$$S_B = \sum_{c=1}^{C} M_c(m_c - m)(m_c - m)^T$$

$$S_W = \sum_{c=1}^{C} \sum_{x \in X_c} (x - m_c)(x - m_c)^T$$

Where the set of training data $X = \{x_1, x_2, \cdots, x_M\} = \bigcup_{c=1}^{C} X_c$ have total C classes, m is the sample mean for the entire data set, $m_c$ is the sample mean for cth class, $M_c = |X_c|$ is the number of samples of a class c and $M = \sum_{c=1}^{C} M_c$. To maximize the between class scatter and minimize the within class scatter the transformation matrix is formulated as

$$W_{opt} = \arg\max_{W} \frac{|W^T S_B W|}{|W^T S_W W|} = [w_1\ w_2\ \cdots\ w_n].$$

PCA is first used with the sample data to reduce its dimensionality. Here, we will call it PCLDA.

### 3.3 Score Fusion Based Weighted Sum Method

One way to combine the scores is to compute a weighted sum as follows:

$$s = \sum w_i s_i$$

where the weight $w_i$ is the amount of confidence we have in the $i$th classifier and its score. In this work, we use 1/EER as a measure of such confidence. Thus, we have a new score

$$s = \sum s_i / EER_i.$$

The weighted sum method based upon the EER is heuristic but intuitively appealing and easy to implement. In addition, this method has the advantage that it is robust to the difference in statistics between training data and test data. Even though the training data and test data have different statistics, the relative strength of the component classifiers is less likely to change significantly, suggesting that the weighting still makes sense. One drawback of the weighed sum method is that the scores generated by component classifiers may have different physical or statistical meaning and different ranges.

### 3.4 Drawbacks Of The Existing System

Most of these methods were initially developed with face images collected under relatively well-controlled conditions and in practice they have difficulty in dealing with the range of appearance variations that commonly occur in unconstrained natural images due to illumination, pose, facial expression, aging, partial occlusions, etc.

In practical face recognition systems, making recognition under uncontrolled lighting condition is important challenge. In Extended Yale-B set the extreme lighting conditions still make it a challenging task for most face recognition methods. The illumination conditions are somewhat less extreme than those of Yale-B, but the induced shadows are substantially sharper, presumably because the angular light sources were less diffuse.

### 3.5 Proposed System

#### 3.5.1 Integral Normalized Gradient Image (INGI)

We can make the following assumptions: 1) most of the intrinsic factor is in the high spatial frequency domain, and 2) most of the extrinsic factor is in the low spatial frequency domain. Considering the first assumption, one might use a high-pass filter to extract the intrinsic factor, but it has been proved that this kind of filter is not robust to illumination variations as shown. In addition, a high-pass filtering operation may have a risk of removing some of the useful intrinsic factor. Hence, we propose an alternative approach, namely, employing a gradient operation. The gradient operation is written as

$$\begin{aligned}\nabla \chi &= \nabla \left( \rho \sum_i \mathbf{n}^T \cdot \mathbf{s}_i \right) \\ &= (\nabla \rho) \sum_i \mathbf{n}^T \cdot \mathbf{s}_i + \rho \nabla \left( \sum_i \mathbf{n}^T \cdot \mathbf{s}_i \right) \\ &\approx (\nabla \rho) \sum_i \mathbf{n}^T \cdot \mathbf{s}_i = (\nabla \rho) W\end{aligned}$$

where the approximation comes from the assumptions that both the surface normal direction (shape) and the light source direction vary slowly across the image, whereas the surface texture varies fast. The scaling factor is the extrinsic factor of our imaging model. The Retinex method and SQI method used the smoothed images as the estimation of this extrinsic factor. We also use the same approach to estimate the extrinsic part

$$\hat{W} = \chi * K$$

where is K a smoothing kernel and denotes the convolution. To overcome the illumination sensitivity, we normalized the gradient map with the following equation:

$$N = \frac{\nabla \chi}{\hat{W}} \approx \frac{(\nabla \rho) W}{\hat{W}} \approx \nabla \rho$$

After the normalization, the texture information in the normalized image $N = \{N_x, N_y\}$ is still not apparent enough. In addition, the division operation may intensify unexpected noise terms. To recover the rich texture and remove the noise at the same time, we integrate the normalized gradients $N_x$ and $N_y$ with the anisotropic diffusion method which we explain in the following, and finally acquire the reconstructed image $\chi_r$.

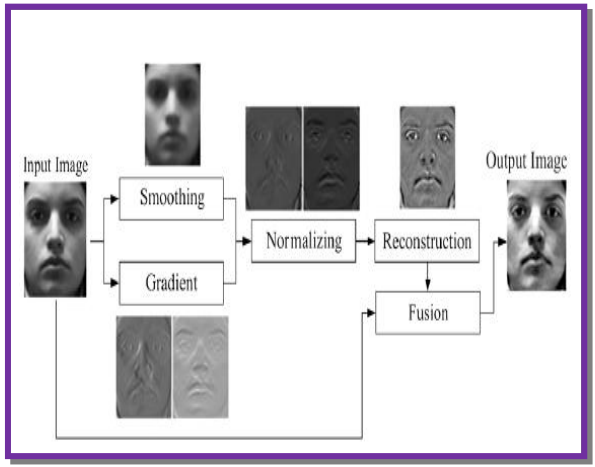

Fig .6 Structure of the integral normalized gradient image

### 3.6 Kernel Principle Component nalysis(KPCA)

KPCA encodes the pattern information based on second order dependencies, i.e., pixel wise covariance among the pixels, and are insensitive to the dependencies of multiple (more than two) pixels in the patterns. Since the eigenvectors in PCA are the orthonormal bases, the principal components are uncorrelated. In other words, the coefficients for one of the axes cannot be linearly represented from the coefficients of the other axes. Higher order dependencies in an image include nonlinear relations among the pixel intensity values, such as the relationships among three or more pixels in an edge or a curve, which can capture important information for recognition. Explicitly mapping the vectors in input space into higher dimensional space is computationally intensive. Using the kernel trick one can compute the higher order statistics using only dot products of the input patterns. Kernel PCA has been applied to face recognition applications and is observed to be able to extract nonlinear features.

The advantage of using KPCA over other nonlinear feature extraction algorithms can be significant computationally. KPCA does not require solving a nonlinear optimization problem which is expensive computationally and the validity of the solution as optimal is typically a concern. KPCA only requires the solution of an eigenvalue problem. This reduces to using linear algebra to perform PCA in an arbitrarily large, possibly infinite dimensional, feature space. The kernel "trick" greatly simplifies calculations in this case. An additional advantage of KPCA is that the number of components does not have to be specified in advance.

KPCA is a useful generalization that can be applied to these domains where nonlinear features require a nonlinear feature extraction tool. We plan to use the KPCA algorithm on real earth science data such as the sea surface temperature (SST) or normalized difference vegetation index (NDVI).The resulting information from KPCA can be correlated with signals such as the Southern Oscillation Index (SOI) for determining relationships with the El Nino phenomenon. KPCA can be used to discover nonlinear correlations in data that may otherwise not be found using standard PCA. The information generated about a data set using KPCA captures nonlinear features of the data. These features correlated with known spatial-temporal signals can discover nonlinear relationships. KPCA offers improved analysis of datasets that have nonlinear structure.

### 3.7 Log-Likelyhood Ratio For Score Fusion

We interpret the set of scores as a feature vector from which we perform the classification task. We have a set of scores $s_1, \ldots, s_n$ computed by classifiers. Now the problem is to decide whether the query-target pair is from the same person

or not based upon these scores. We can cast this problem as the following hypothesis testing:

$$H_0 : S_1,\ldots,S_n \sim p(s_1,\ldots,s_n|\text{diff})$$
$$H_1 : S_1,\ldots,S_n \sim p(s_1,\ldots,s_n|\text{same})$$

where $p(s_1,\ldots,s_n|\text{diff})$ is the distribution of the scores when the query and target are from different persons, and $p(s_1,\ldots,s_n|\text{same})$ is the distribution of the scores when the query and target are from the same person. The figure gives example of such distributions and provides the intuition behind the benefits of using multiple scores generated by multiple classifiers. Suppose we have two classifiers and they produce two scores $s_1$ and $s_2$. and Figure shows $p(s_1|\text{diff})$ and $p(s_1|\text{same})$, distributions of a single score.The region between the two vertical lines is where the two distributions overlap. Intuitively speaking, a classification error can occur when the score falls within this overlapped region, and the smaller this overlapped region, the smaller the probability of classification error. Likewise, Fig. 9(b) shows $p(s_2|\text{diff})$ and $p(s_2|\text{same})$. Fig(c) shows how the pair of the two scores $(s_1, s_2)$ is distributed. The upper left of Fig(c) is the scatter plot of $(s_1, s_2)$ when the query and target are from the same person, and the upper right of Fig. 9(c) is the scatter plot of when the query and target are from different persons. The bottom of Fig(c) shows how the two scatter plots overlap.

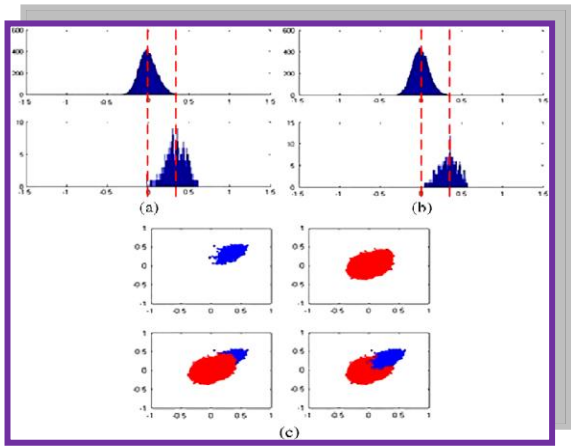

Fig .7 Score distribution of two classifiers

Compared with Fig. (a) and (b), we can see that the probability of overlap can be reduced by jointly considering the two scores $s_1$ and $s_2$, and , which suggests that hypothesis testing based upon the two scores $s_1$ and $s_2$ is better than hypothesis testing based upon a single score $s_1$ or $s_2$.

If we know the two densities $p(s_1,\ldots,s_n|\text{diff})$ and $p(s_1,\ldots,s_n|\text{same})$, the log-likelihood ratio test achieves the highest verification rate for a given false accept rate

$$\log \frac{p(s_1,\ldots,s_n|\text{same})}{p(s_1,\ldots,s_n|\text{diff})} \gtrless 0.$$

However, the true densities $p(s_1,\ldots,s_n|\text{diff})$ and $p(s_1,\ldots,s_n|\text{same})$ are unknown, so we need to estimate these densities observing scores computed from query-target pairs in the training data. One way to estimate these densities is to use a nonparametric density estimation. In this work, we use parametric density estimation in order to avoid over-fitting and reduce computational complexity. In particular, we model the distribution of $s_i$ given $H_0$ as a Gaussian random variable with mean $m_{\text{diff},i}$ and variance $\sigma^2_{\text{diff},i}$, and model $\{S_i\}_{i=1}^n$ given $H_0$ as independent Gaussian random variables with density

$$p(s_1,\ldots,s_n|\text{diff}) = \prod N\left(s_i; m_{\text{diff},i}, \sigma^2_{\text{diff},i}\right)$$

$$N(x; m, \sigma^2) = (1/\sqrt{2\pi\sigma^2})\exp\left(-((x-m)^2/2\sigma^2)\right)$$

Where is the Gaussian density function. The parameters $(m_{\text{diff},i}, \sigma^2_{\text{diff},i})$ are estimated from the scores of the $i$th classifier corresponding to nonmatch query-target pairs in the training database. Similarly, we approximate the density of given $H_1$ by $\prod N(s_i; m_{\text{same},i}, \sigma^2_{\text{same},i})$, and the parameters $\{S_i\}_{i=1}^n$ $(m_{\text{diff},i}, \sigma^2_{\text{diff},i})$ and $(m_{\text{same},i}, \sigma^2_{\text{same},i})$ are computed from the scores of the classifier corresponding to match query-target pairs in the training database. Now we define the fused score to be the log-likelihood ratio, which is given by

$$S = \log \frac{N\left(S_i; m_{\text{same},i}, \sigma^2_{\text{same},i}\right)}{N\left(S_i; m_{\text{diff},i}, \sigma^2_{\text{diff},i}\right)}$$
$$= \sum_{i=1}^n \left[\frac{(S_i - m_{\text{diff},i})^2}{2\sigma^2_{\text{diff},i}} - \frac{(S_i - m_{\text{same},i})^2}{2\sigma^2_{\text{same},i}}\right] + c$$

## 4. Conclusion

In the proposed approach we have through a whole face recognition system**.** The Retinex method and SQI method are used to smoothed images as the estimation of the extrinsic factor. KPCA offers improved analysis of datasets that have nonlinear structure. It is also useful for generalization that can be applied to the domains where nonlinear features are required. The overall process can be viewed as a pipeline consisting of image normalization, feature extraction, and subspace representation, each stage increases resistance to illumination variations and makes the information needed for recognition more manifest. This method achieves very significant improvements, than the other method of verification rate is 88.1% at 0.1% false acceptance rate. The integrative framework is proposed to verify the face in various lightening conditions.

## 5. Experimental Results

The proposed work narrates the frame work of the experimental result as we take the gray scale image (0,255) as the original image. Fig 8 shows the Test image which is to be preprocessed. Fig 9 illustrates the preprocessed image 1 Fig.10 shows the image as we stored in the data base. Fig .11 demonstrates the final preprocessed image3.

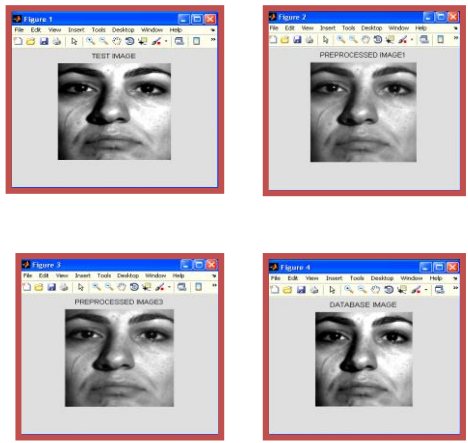

## 6. REFERENCES

[1] P. J. Phillips, H. Moon, S. A. Rizvi, and P. J. Rauss, "The FERET evaluation methodology for face recognition algorithms," *IEEE Trans. Pattern Anal. Mach. Intell.*, vol. 22, no. 10, pp. 1090–1104, Oct. 2000.

[2] D. M. Blackburn, M. Bone, and P. J. Phillips, "Facial recognition vendor test 2000 evaluation report," Dec. 2000 [Online]. Available: http://www.frvt.org/

[3] P. Phillips, P. Grother, R. Micheals, D. Blackburn, E. Tabassi, and M. Bone, "Face recognition vendor test 2002: evaluation report," 2003 [Online]. Available: http://www.frvt.org/

[4] K. Messer, J. Kittler, M. Sadeghi, M. Hamouz, and A. Kostin *et al.*, "Face authentication test on the BANCA database," in *Proc. Int. Conf. Pattern Recognit.*, Aug. 2004, vol. 4, pp. 523–532.

[5] P. J. Phillips, P. J. Flynn, T. Scruggs, K. Bowyer, J. Chang, K. Hoffman,
J. Marques, J. Min, and W. Worek, "Overview of the face recognition grand challenge," in *Proc. IEEE. Comput. Vis. Pattern Recognit.*, Jun.
2005, vol. 1, pp. 947–954.

[6] P. N. Belhumeur and D. J. Kriegman, "What is the set of images of an object under all possible lighting conditions?," in *Proc. IEEE Conf. Comput. Vis. Pattern Recognit.*, Jun. 1996, pp. 270–277.

[7] R. Ramamoorthi and P. Hanrahan, "On the relationship between radiance and irradiance: Determining the illumination from images of a convex Lambertian object," *J. Opt. Soc. Amer.*, vol. 18, no. 10, pp. 2448–2459, 2001.

[8] A. Shashua and T. Riklin-Raviv, "The quotient image: Class-based re-rendering and recognition with varying illuminations," *IEEE Trans.
Pattern Anal. Mach. Intell.*, vol. 23, no. 2, pp. 129–139, Feb. 2001.

[9] H. Wang, S. Li, and Y. Wang, "Generalized quotient image," in *Proc. IEEE. Comput. Vis. Pattern Recognit.*, Jul. 2004, vol. 2, pp. 498–505.

[10] Q. Li, W. Yin, and Z. Deng, "Image-based face illumination transferring using logarithmic total variation models," *Int. J. Comput. Graph.*, vol. 26, no. 1, pp. 41–49, Nov. 2009.

[11] E. H. Land, "The Retinex theory of color vision," *Sci. Amer.*, vol. 237, no. 6, pp. 108–128, Dec. 1977.

[12] D. J. Jobson, Z. Rahman, and G. A. Woodell, "Properties and performance
of a center/surround Retinex," *IEEE Trans. Image Process.*, vol. 6, no. 3, pp. 451–462, Mar. 1997.

[13] R. Gross and V. Brajovie, "An image preprocessing algorithm for illumination
invariant face recognition," in *Proc. 4th Int. Conf. Audio Video Based Biometric Person Authentication*, 2003, vol. 2688/2003,
pp. 10–18.

[14] J. Malik and P. Perona, "Scale-space and edge detection using anisotropic diffusion," *IEEE Trans. Pattern Anal. Mach. Intell.*, vol. 12, no. 7, pp. 629–639, Jul. 1990.

[15] M. A. Turk and A. P. Pentland, "Eigenfaces for recognition," *J. Cogn.
Neurosci.*, vol. 3, no. 1, pp. 71–86, 1991.

[16] P. S. Penev and J. J. Atick, "Local feature analysis: A general statistical
theory for object representation," *Network: Comput. Neural Syst.*, vol. 7, no. 3, pp. 477–500, 1996.